\documentclass{article}
\usepackage{graphicx}
\usepackage{graphics}
\usepackage{amsfonts}
\usepackage{marvosym}
\newcommand{\R}{\mathbb{R}}

\begin{document}
\title{From Data to the p-Adic or Ultrametric Model}

\author{Fionn Murtagh  \\
Science Foundation Ireland, Wilton Place, Dublin 2, Ireland, and  \\
Department of Computer Science, Royal Holloway, University of 
London \\ Egham TW20 0EX, England \\
Email: fmurtagh@acm.org}

\maketitle

\begin{abstract}
We model anomaly and change in data by embedding the data in an
ultrametric space.  Taking our initial data as cross-tabulation 
counts (or other input data formats), Correspondence Analysis allows
us to endow the information space with a Euclidean metric. We then 
model anomaly or change by an induced ultrametric.  
The induced ultrametric that we are particularly interested in 
takes a sequential -- e.g.\ temporal -- ordering of the data into 
account.  We apply this work to the flow of narrative expressed in 
the film script of the Casablanca movie; and to the evolution between 
1988 and 2004 of the Colombian social conflict and violence.   
\end{abstract}
\maketitle

\section{Modeling of Anomaly or Change: Introduction}

The data mining and data analysis challenges addressed are the 
following. 
(i) Great masses of data, textual and otherwise, need to be 
exploited and decisions need to be made.  Correspondence Analysis 
handles multivariate numerical and symbolic data with ease. 
(ii) Structures and interrelationships evolve in time.
(iii) We must consider a complex web of relationships.
(iv) We need to address all these issues from data sets and data flows.   

Various aspects of how we respond to these challenges will be discussed 
in this article, complemented by the Appendix.   
We will look at how this works, using the Casablanca film script,
and data from the long Colombian civil strife involving government, guerrillas,
paramilitaries and civilians. 

\section{The Geometry and Topology of Information}

We consider Correspondence Analysis and hierarchical clustering as a 
semantic analysis platform.  To illustrate our description, we will 
take film script, the semi-structured expression of a story.  Film script
is the starting point of what may become a movie.  

For McKee \cite{mckee},  film script 
text is the ``sensory surface of a work of 
art'' and reflects the underlying emotion or perception.  Our data
mining approach models and tracks these underlying aspects in 
the data.  Our approach to textual data mining has a range of novel elements.
  

The starting point for analysis 
is frequency of occurrence data, 
typically the ordered scenes crossed by all words used in the script.  

If the totality of interrelationships is one facet of semantics, 
then another is anomaly (or change, novelty, breakpoint) 
as modeled by a clustering 
hierarchy.  If, therefore, a scene is quite different from immediately 
previous scenes, then it will be incorporated into the hierarchy at a 
high level.  This novel view of hierarchy will be discussed further 
in section \ref{sectgeotop} below.   

We draw on these two vantage points on semantics -- 
viz.\ totality of interrelationships, and using a hierarchy to express change.
See \cite{khrenn} for other work that uses p-adic metric properties,  
tantamount to ultrametric properties, for the same goal of change detection.  

\subsection{Modeling Semantics via the Geometry and Topology of 
Information}
\label{sectgeotop}

Some underlying principles are as follows.  
We start with the cross-tabulation data, scenes $\times$ attributes.
Scenes and attributes are embedded in a metric space.
This is how we are probing the {\em geometry of information},
which is a term and viewpoint used by \cite{vanr}.  

\begin{figure}
\begin{center}
\includegraphics[width=6cm]{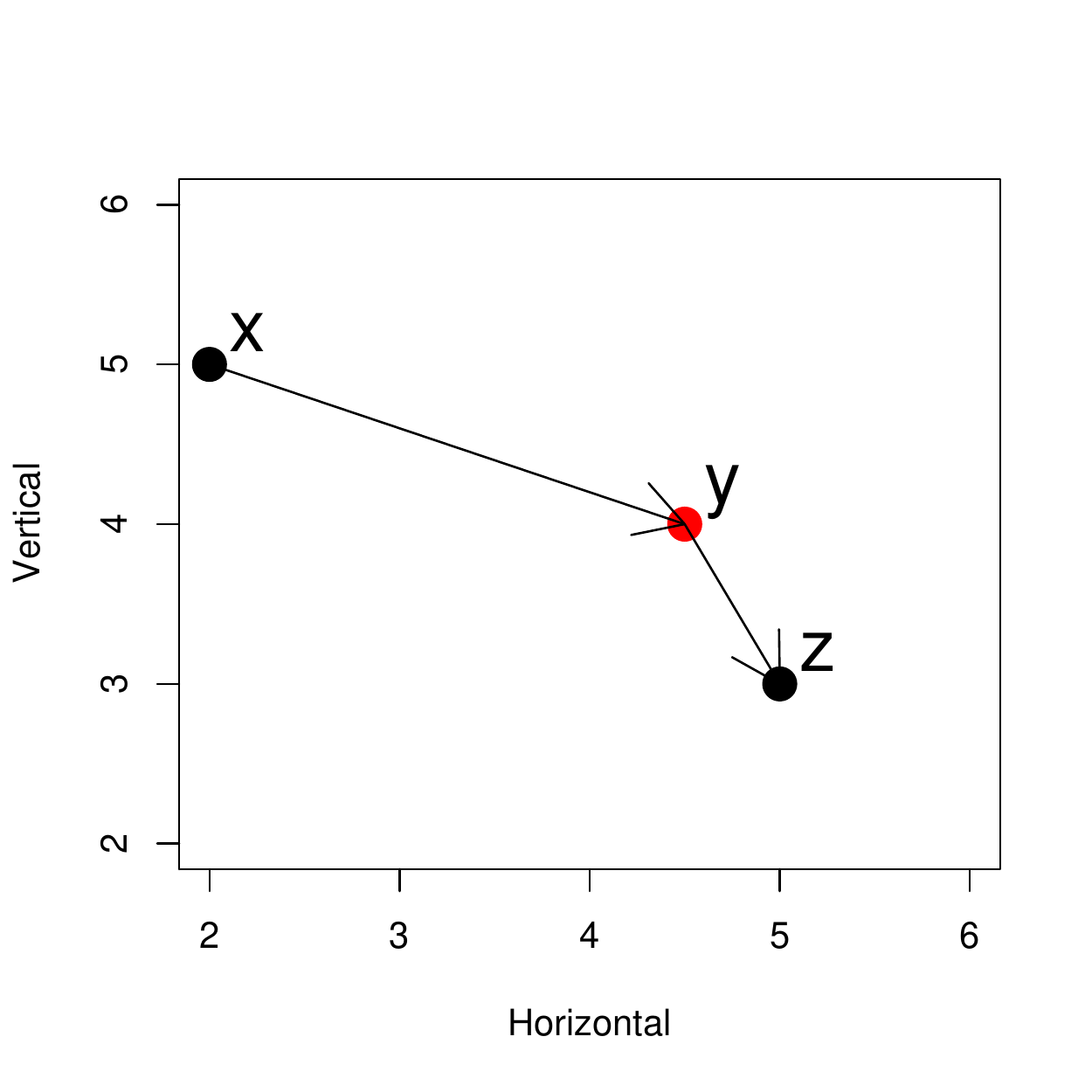}
\end{center}
\caption{The triangular inequality defines a metric: 
every triplet of points satisfies the relationship:
$d(x,z) \leq d(x,y) + d(y,z)$ for distance $d$.}
\label{fig2}
\end{figure}

Underpinning the display in Figure \ref{fig1} is a Euclidean embedding.
The triangular inequality holds for metrics.  An example of a metric is 
the Euclidean distance, exemplified in Figure \ref{fig2}, where each 
and every triplet of points satisfies the relationship:
$d(x,z) \leq d(x,y) + d(y,z)$ for distance $d$.  Two other relationships
also must hold.  These are symmetry and positive definiteness, respectively: 
$d(x,y) = d(y,x)$, and $d(x,y) > 0 $ if $x \neq y$, $d(x,y) = 0 $ if $x = y$.

Further underlying principles used in Figure \ref{fig1} are as follows.
The axes are the principal axes of momentum.
Identical principles are used as in classical mechanics.
The scenes are located as weighted averages of all associated 
attributes; and vice versa. 

Huyghens' theorem relates to decomposition of 
inertia of a cloud of points. This is the basis of Correspondence Analysis.

We come now to a different principle: that of the {\em topology of 
information}.   The particular topology used is that of hierarchy.
Euclidean embedding provides a very good starting point to look at 
hierarchical relationships.
An innovation in our work is as follows: the hierarchy takes sequence, e.g.\
timeline, into account.
This captures, in a more easily understood way, the notions of
novelty, anomaly or change.

\begin{figure}[t]
\begin{center}
\includegraphics[width=6cm]{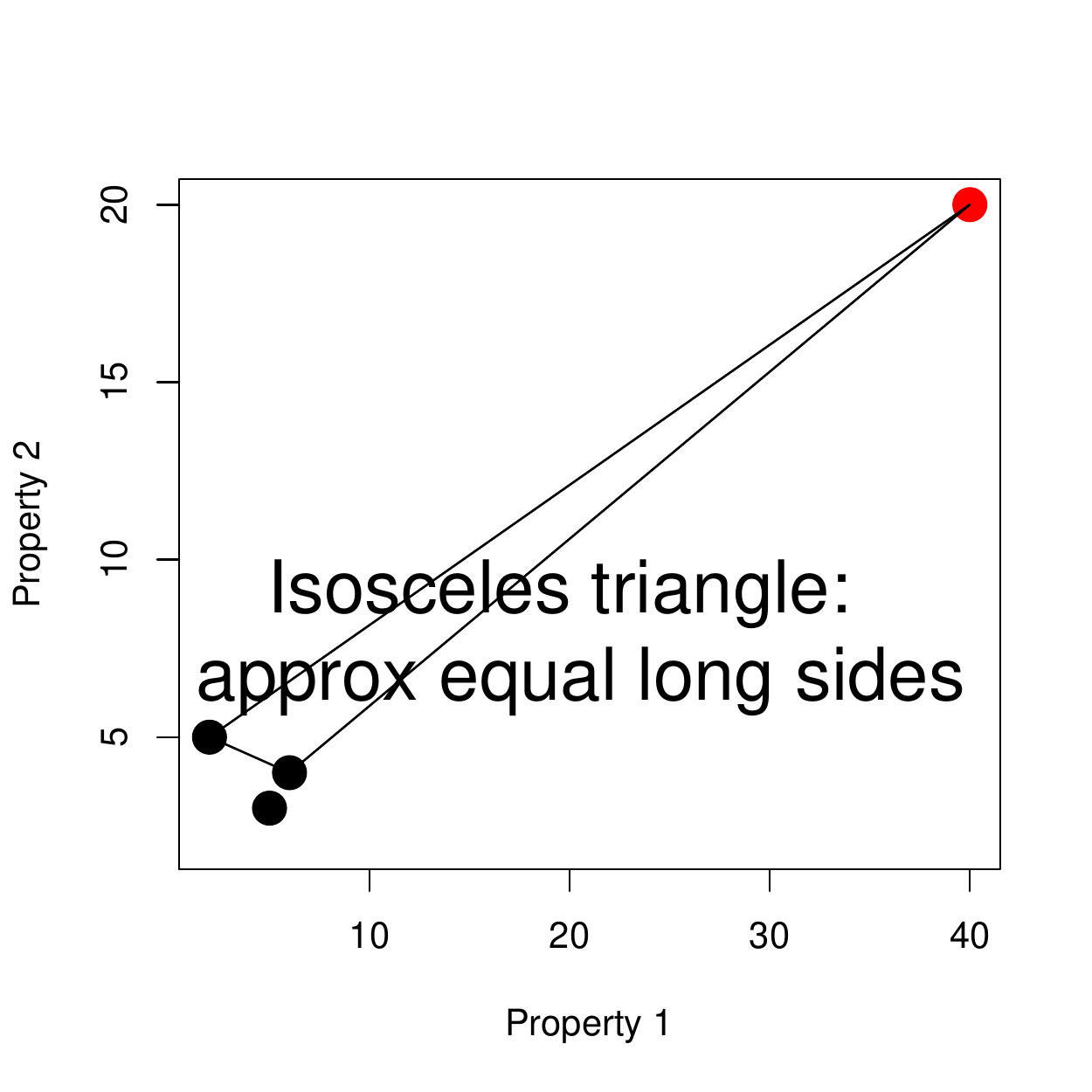}
\includegraphics[width=6cm]{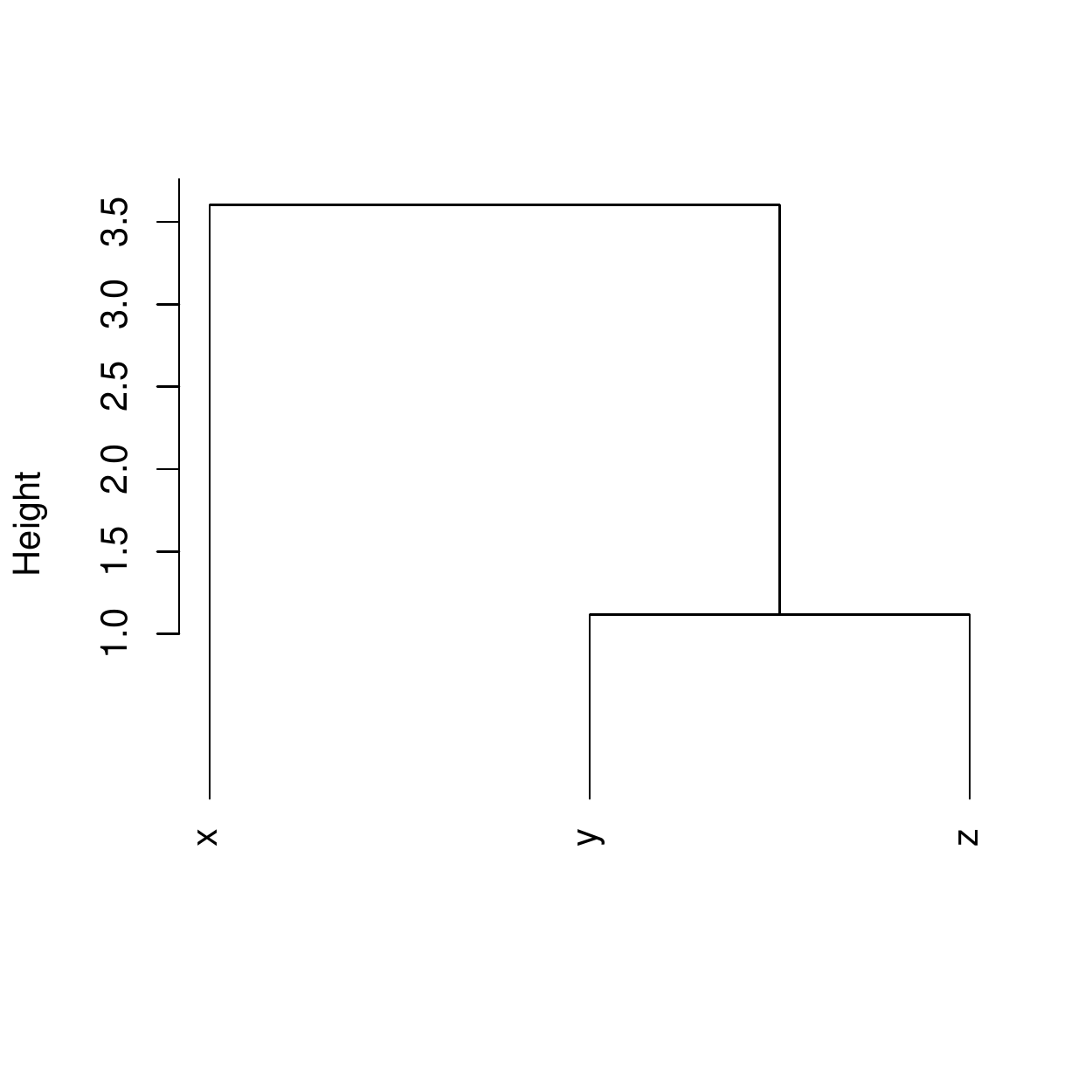}
\end{center}
\caption{Left: The query is on the far right.  While we can easily 
determine the closest target (among the three objects represented
by the dots on the left), is the closest really that much different
from the alternatives?
Right: The strong triangular inequality defines an ultrametric: 
every triplet of points satisfies the relationship:
$d(x,z) \leq \mbox{max} \{ d(x,y), d(y,z) \}$ for distance $d$.
Cf.\ by reading off the hierarchy, how this is verified for all $x, y, 
z$: $d(x,z) = 3.5; d(x,y) = 3.5; d(y,z) = 1.0$.   
In addition the symmetry and positive definiteness conditions 
hold for any pair of points.}
\label{fig4}
\end{figure}

Let us take an informal case study to see how this works.  
Consider the situation of seeking documents based on titles.  If the target
population has at least one document that is close to the query, 
then this is (let us assume) clearcut.  However if all documents in the 
target population are very unlike the query, does it make any sense to 
choose the closest?  Whatever the answer here we are focusing on the 
inherent ambiguity, which we will note or record in an appropriate way.  
Figure \ref{fig4}, left, illustrates this situation where
the query is the point to the right.   

By using approximate similarity this situation can be 
modeled as an isosceles triangle with small base, as illustrated in
Figure \ref{fig4}, left.  
An ultrametric space has properties that are very unlike a metric
space, and one such property is that the only triangles allowed are either
(i) equilateral, or (ii) isosceles with small base.  So Figure \ref{fig4}
can be taken as representing a case of ultrametricity.  What this means
is that the query can be viewed as having a particular sort of dominance
or hierarchical relationship vis-\`a-vis any pair of target documents.
Hence any triplet of points here, one of which is the query (defining the 
apex of the isosceles, with small base, triangle), defines local hierarchical
or ultrametric structure.  (See \cite{murtagh04} for case studies.) 

It is clear from Figure \ref{fig4} that we should use approximate
equality of the long sides of the triangle.  The further away the query is 
from the other data then the better is this approximation \cite{murtagh04}.

What sort of explanation does this provide for our conundrum?  It means that 
the query is a novel, or anomalous, or unusual ``document''.   It is up to 
us to decide how to treat such new, innovative cases.  It raises though the 
interesting perspective that here we have a way to model and
subsequently handle the semantics of anomaly or innocuousness.  

\section{The Changing Nature of Movie and Drama}

\subsection{Background}

McKee \cite{mckee} bears out the great importance of the 
film script: ``50\% of what we understand comes from watching it being 
said.''  And: ``A screenplay waits for the camera. ... Ninety percent 
of all verbal expression has no filmic equivalent.''

An episode of a television series costs US\$ 2--3 million per 
one hour of television.  
Generally screenplays are written speculatively or commissioned, and then 
prototyped by the full production of a pilot episode.  Increasingly, and
especially availed of by the young, television series are delivered via the 
Internet.  Originating in one medium -- cinema, television, game, online
-- film and drama series are increasingly migrated to another.   
So scriptwriting must take account of digital multimedia platforms.  
This has been referred to in computer networking parlance as ``multiplay''
and in the television media sector as a ``360 degree'' environment.  

Cross-platform delivery motivates interactivity in drama.  So-called
reality TV has a considerable degree of interactivity, as well as being
largely unscripted.  

There is a burgeoning need for us to be in a position to model the
semantics of film script, -- its most revealing structures, patterns
and  layers.
With the drive towards interactivity, we also want to leverage 
this work towards more general scenario analysis.  Other potential 
applications are to business strategy and planning; education 
and training; and science, technology and economic development policy.

\subsection{Casablanca Narrative: Illustrative Analysis}

The well known Casablanca movie serves as an example for us.  
Film scripts, such as for Casablanca, are partially structured
texts.  Each scene has metadata and the body of the scene contains
 dialog and possibly other descriptive data.  
The Casablanca script was half completed when production began in 
1942.  The dialog for some scenes was written while shooting was in 
progress.  Casablanca was based on an unpublished 1940 screenplay 
\cite{burnett}.  It was scripted by J.J. Epstein, P.G. 
Epstein and H. Koch.  The film was 
directed by M. Curtiz and produced by H.B. Wallis and J.L. Warner.
It was shot by Warner Bros.\ between May and August 1942.

A data set was constructed from the 77 successive scenes crossed by 
attributes -- Int[erior], Ext[erior], Day, Night, Rick, Ilsa, Renault, 
Strasser, Laszlo, Other (i.e.\ minor character), and 29 locations.  
Many locations were met with just once; Rick's Caf\'e was the 
location of 36 scenes.  In scenes based in Rick's Caf\'e we did 
not distinguish between ``Main 
room'', ``Office'', ``Balcony'', etc.  Because of the plethora of 
scenes other than Rick's Caf\'e we assimilate these to just one, ``other than
Rick's Caf\'e'', scene.

\begin{figure}[t]
\begin{center}
\includegraphics[width=16cm]{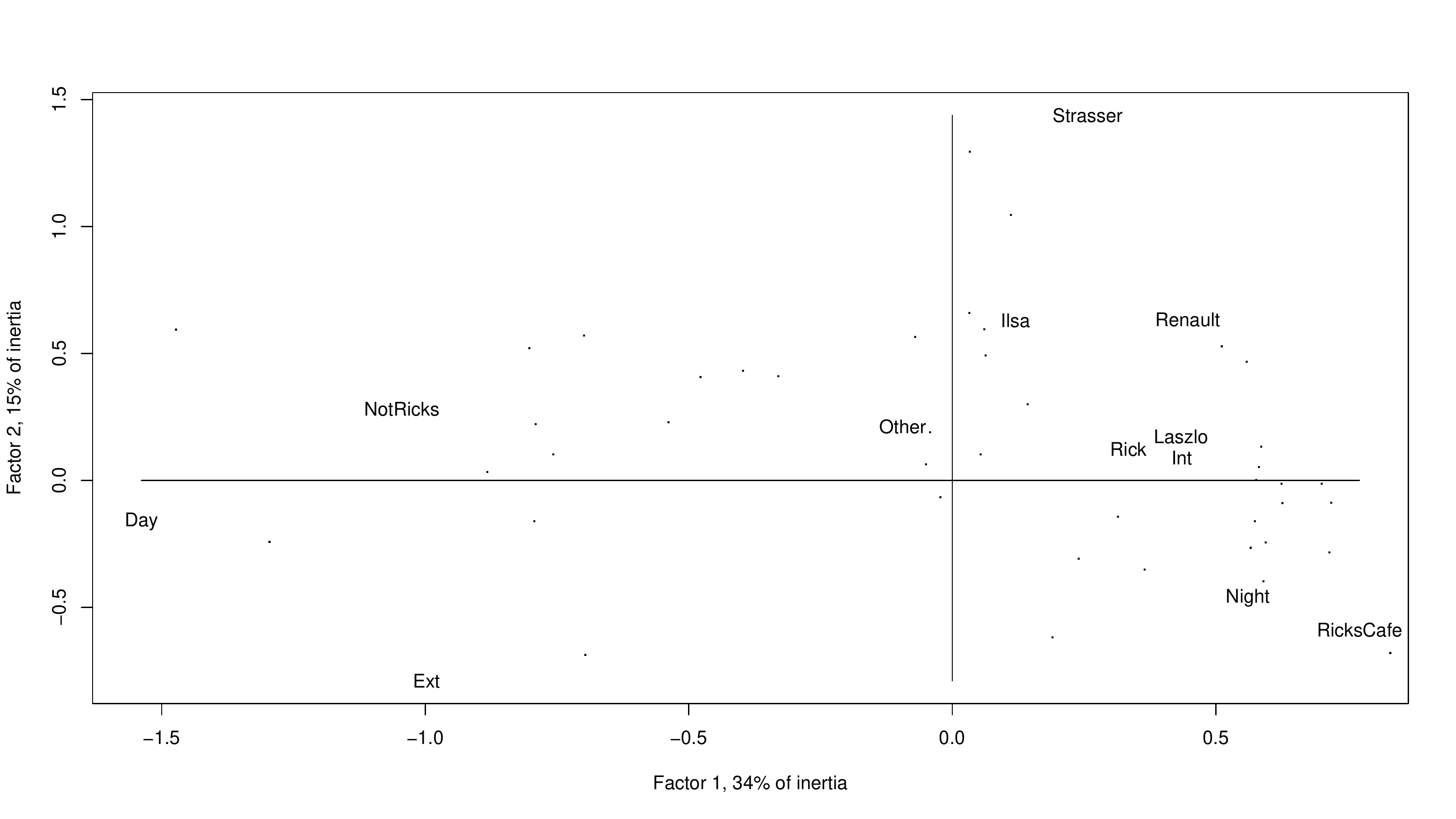}
\end{center}
\caption{Correspondence Analysis of the Casablanca data derived from
the script.  The input data is presences/absences for 77 scenes
crossed by 12 attributes.  The 77 scenes are located at the dots, 
which are not labeled here for clarity.  For a short review of the 
analysis methodology, see Appendix.}
\label{fig1}
\end{figure}

In Figure \ref{fig1}, 12 attributes are displayed; 77 scenes are 
displayed as dots (to avoid over-crowding of labels).  
Approximately 34\% (for factor 1) + 15\% (for factor 2) = 
49\% of all information, expressed as 
inertia explained, is displayed here. 
We can study interrelationships between characters, other attributes, 
scenes, for instance closeness of Rick's Caf\'e with Night and 
Int (obviously enough).  

Figure \ref{fig6} uses a sequence-constrained complete link agglomerative
algorithm.  It shows up scenes 9 to 10, and progressing from 39, to 40 and 41,
as major changes.  The sequence constrained algorithm, 
i.e.\ agglomerations are 
permitted between adjacent segments of scenes only, is described in an 
Appendix to this article, and in greater detail in \cite{murtagh05}.
The agglomerative criterion used, that is subject to this sequence constraint, 
is a complete link one.  

A study in greater depth of the semantics of Casablanca can be found in 
\cite{murtaghetal08}. We refer the reader to that work for further reading
since, there, we use all words appearing in the screenplay rather than the 
set of 12 attributes that we limit ourselves to here.  

\begin{figure}
\begin{center}
\includegraphics[width=18cm,angle=-90]{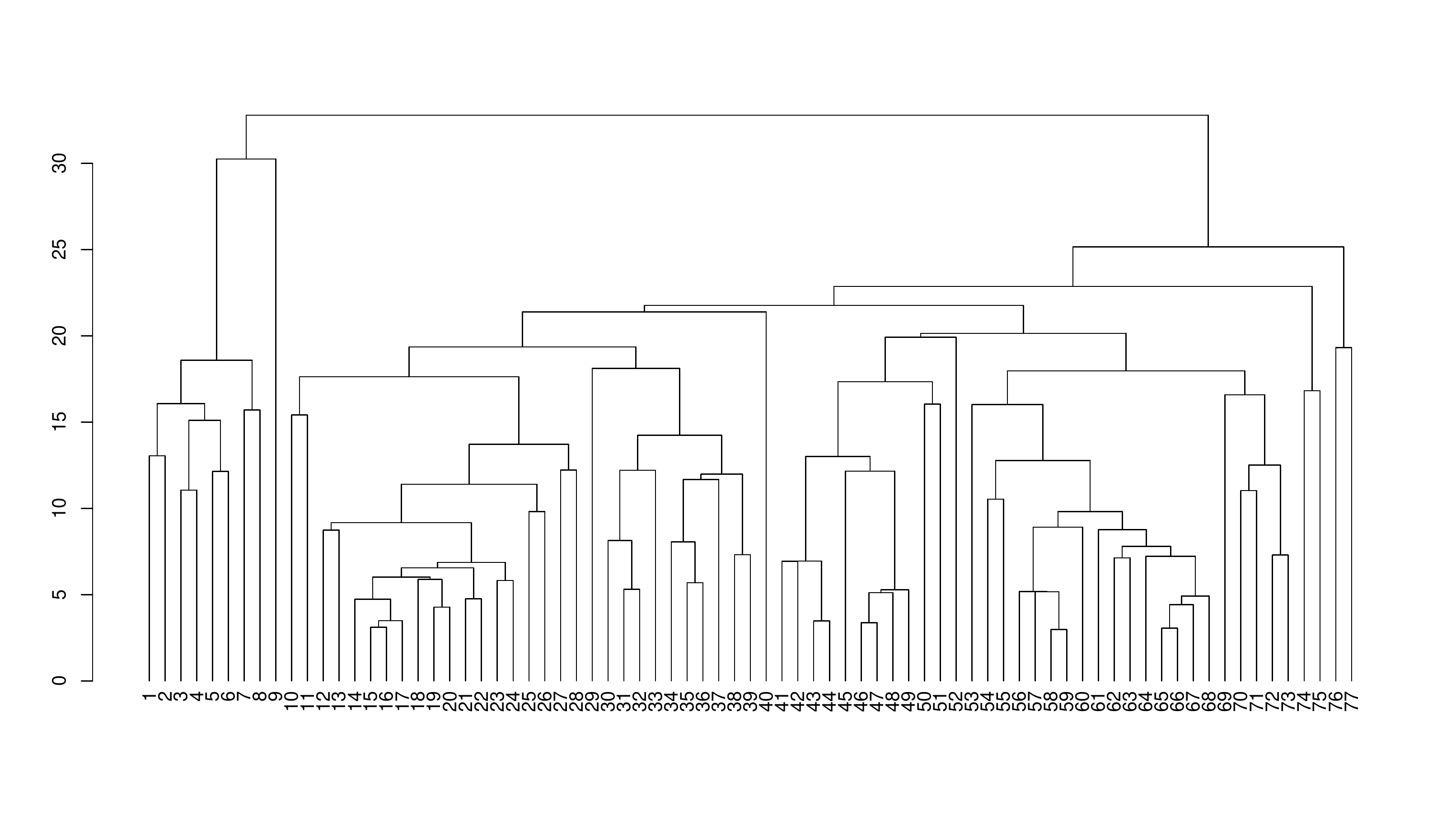}
\end{center}
\caption{77 scenes clustered.  These scenes are in sequence: a 
sequence-constrained agglomerative criterion is used for this.  The 
agglomerative criterion itself is a complete link one.  See \cite{murtagh85}
for properties of this algorithm.}
\label{fig6}
\end{figure}

\subsection{Our Platform for Analysis of Semantics}

Correspondence analysis supports the following:
analysis of multivariate, mixed numerical/symbolic data;
web of interrelationships; and 
evolution of relationships over time.

Correspondence Analysis is in practice {\em a tale of three metrics}
\cite{murtagh05}.  The analysis is based on embedding 
a cloud of points from a space 
governed by one metric into another.  Furthermore the cloud offers 
vantage points of both observables and their characterizations, so 
-- in the case of film script -- for 
any one of the metrics we can effortlessly pass between the space of 
film script scenes and attribute set.  The three metrics are as follows. 

\begin{itemize}
\item Chi squared, $\chi^2$, metric -- 
appropriate for profiles of frequencies of occurrence.
\item Euclidean metric, for visualization, and for static context.
\item 
Ultrametric, for hierarchic relations and, as we use it in this work,
for dynamic context.
\end{itemize}

In the analysis of semantics, we distinguish two separate aspects.

Firstly there is context -- the collection of all interrelationships. 
The Euclidean distance makes a lot of sense when the 
population is homogeneous. All interrelationships together provide context, 
relativities -- and hence meaning.  

Secondly there is hierarchy which tracks anomaly.
Ultrametric distance makes a lot of sense when the 
observables are heterogeneous, discontinuous.
The latter is especially useful for determining anomalous (or 
atypical or innovative) cases.

\section{Colombia Social Conflict: Analysis of Change}

\subsection{Background of the Data}
\label{sect51}

The data used in this work related to social conflict and were produced
by CERAC, the Conflict Analysis Resource Center, www.cerac.org.co.   We will
refer to the data used as the CERAC Colobmia Conflict database.  It is
being grown on an ongoing basis.

From \cite{rsv1}, we use high frequency micro-data, relating to
internal social conflict in Colombia (population: 44 million) from 1988
to 2004, hence over a period of 17 years.  Social conflict is
not ethnic, religious, nor regional, as often the case elsewhere,
but is instead economical, political, and military, in origin.  Economic
factors, for instance, include the narcotics sector and kidnapping.
A short periodization of the conflict
is as follows \cite{rsv1}: 
1988--1991, ``adjustment period'' due to the end of the Cold War; 
1992--1996, ``stagnation period''; and from 1997 onwards, 
``upsurge period''.

The CERAC data relates to
actions, and their intensity, leading to information on impact.
Armed combat is taken therefore as a clash, or an attack (relating,
respectively, to bilateral or multilateral, versus unilateral, engagement).
For the data period, the Colombian conflict has seen more than
3000 casualties per year. 
\cite{rsv2} discusses the effect of Alvaro Uribe becoming President in August
2002.

The study \cite{rsv3} underlines interest in civil war dynamics:
``... We consider ... different attack types (unopposed events) plus
clashes ... For each event type we determine the number of civilian
killings and injuries ... and the population density ... We argue that
policy must focus on ...  very specific circumstances for civilian
casualties ... [such as] massacres by illegal right-wing paramilitaries in
rural areas --- ... These events account for almost 40\% of all conflict
casualties ...''.

In this article, 
we study change versus continuity over time, allowing for gradation
in such change.  A hierarchical clustering is used, taking account of
the timeline, furnishing a visualization of 
breakpoints and timeline resolution-related properties.

\subsection{Correspondence Analysis and Metric Embedding}

Starting with an array of counts of presence versus absence, or frequency of
occurrence, which provides data that cross-tabulates a set of observations
and a set of attributes,  we can embed the observations and attributes in a
Euclidean space.  This factor space is mathematically optimal in a certain
sense (using the least squares criterion, which is also Huyghens' principle of
decomposition of inertia).  Furthermore a
Euclidean space allows for easy visualization that would be more awkward
to bring about otherwise.  A third reason for embedding the observations and
attributes in a Euclidean factor space is that weighting of observations
and attributes is handled naturally in this framework.

We used 144 numerical attributes.  These related to numbers
killed or otherwise effected.   There were 20,288 successive
events.

Aggregating by month the numerical data, 144 attributes used, yielded
data for 204 successive months.   The embedding in a Euclidean factor
space using Correspondence Analysis was 80-dimensional.  Hierarchical
clustering of the 204 months, using this full-dimensional 
Euclidean embedding, gave the
hierarchy shown in Figure \ref{fig20}.  This hierarchy is built on
successive aggregated months.  Hence it is an adjacency-respecting, or
contiguity-constrained, hierarchical clustering.  
As before, the contiguity-constrained complete link
agglomerative hierarchical clustering method is used.

\subsection{The Hierarchy of Changepoints}

\begin{figure}
\includegraphics[width=18cm,angle=-90]{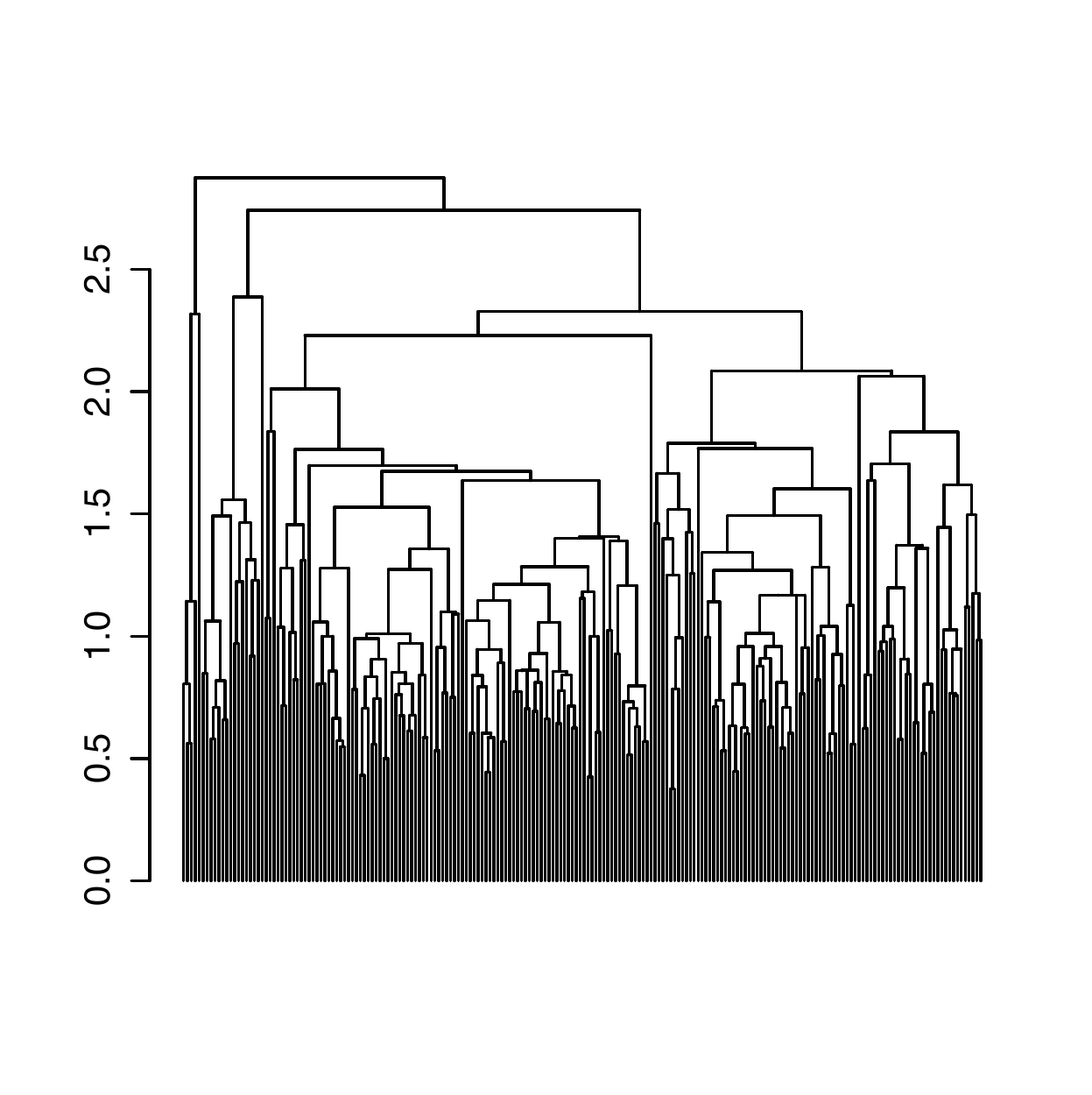}
\caption{Hierarchical clustering of the 204 successive months, based on the
144 numerical attributes.  Leaf nodes are in sequence from month 1 (top here)
to month 204 (bottom here).}
\label{fig20}
\end{figure}

In Figure \ref{fig20} changepoints of some significance are to be found
between months 21 and 22; and 119, 120 and 121.  Very early on, there is a
clear changepoint between months 5 and 6.  A less sizable, but
nonetheless clear, changepoint
is to be found between months 172 and 173.  If we label the clusters of 
the 8-cluster partition as 1 through 8, we find the partition corresponding
to that horizontal cut of the tree of Figure \ref{fig20} to be, for the 
204 months in sequence: 

\begin{verbatim}
1 1 1 1 2 3 3 3 3 3 3 3 3 3 3 3 3 3 3 3 4 5 5 5 5 5 5 5 5 5 5 5 5 5 
5 5 5 5 5 5 5 5 5 5 5 5 5 5 5 5 5 5 5 5 5 5 5 5 5 5 5 5 5 5 5 5 5 5 
5 5 5 5 5 5 5 5 5 5 5 5 5 5 5 5 5 5 5 5 5 5 5 5 5 5 5 5 5 5 5 5 5 5 
5 5 5 5 5 5 5 5 5 5 5 5 5 5 5 5 5 6 7 7 7 7 7 7 7 7 7 7 7 7 7 7 7 7 
7 7 7 7 7 7 7 7 7 7 7 7 7 7 7 7 7 7 7 7 7 7 7 7 7 7 7 7 7 7 7 7 7 7 
7 7 8 8 8 8 8 8 8 8 8 8 8 8 8 8 8 8 8 8 8 8 8 8 8 8 8 8 8 8 8 8 8 8
\end{verbatim}

\begin{figure}[t]
\includegraphics[width=8cm]{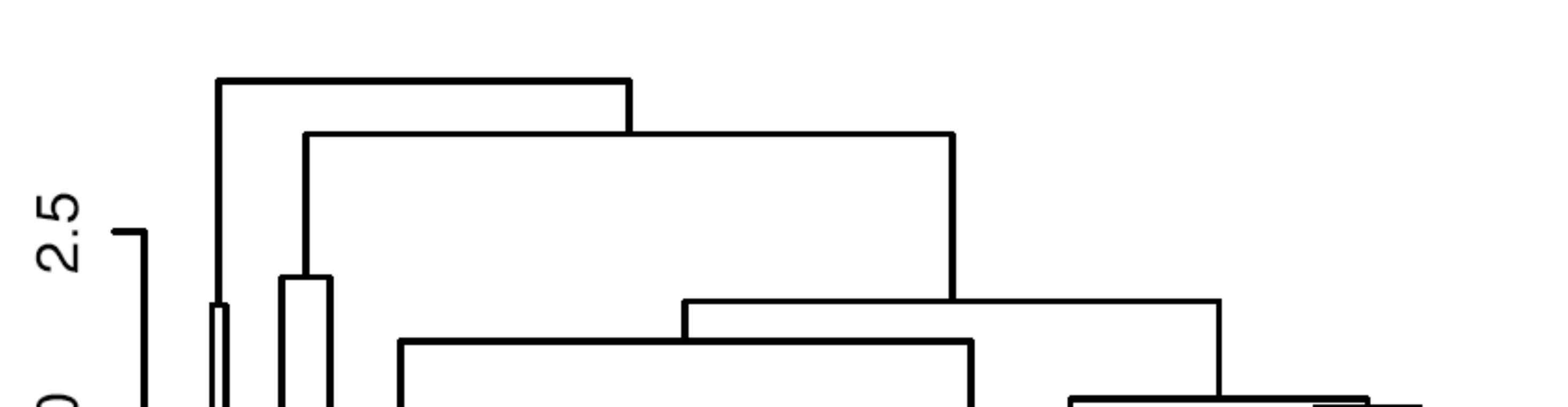}
\caption{From Figure \ref{fig20}, the top of the hierarchy is shown, 
based on a partition with 8 clusters.  Clusters 2, 4 and 6, here have 
cardinalities of just one.  As discussed in the text, cluster 1 
comprises months 1 to 4.  Cluster 3 is months 6 to 20.  Cluster 5 is 
months 22 to 119.  Cluster 7 is months 121 to 172.  Finally cluster 8 
is months 173 to 204.}
\label{fig30}
\end{figure}

We will look closely at the changes between months (i) 21 and 22; 
(ii) 119, 120 and 
121; and (iii) 172 and 173.  From Figure \ref{fig30}, these 
correspond to the caesuras in moving through periods defined by
(i) clusters of months 3, 4 and 5; (ii) clusters of months 5, 6 and 7; 
and (iii) clusters of months 7 and 8.

Since the 204 months are from January 1988 to December 
2004, we determine these changes as being, respectively: (i) late (September, 
October) 1989; (ii) beginning of 1998 (December 1997, January and February 1998); 
and (iii) early 2002 (April, May 2002).  

The 1989 changepoint, (i), as noted above (section \ref{sect51} and 
\cite{rsv1}) was in the ``adjustment period'' due to the end of the Cold War
and its influence on the Colombian civil strife.  

The 1998 changepoint, (ii), was in a period when guerrilla gains from 1996 
onwards were being reversed, a process that was seen to be so by 1999.  
There were high levels of government casualties in 1997--1998, due to 
guerrilla operations against isolated military and police bases.  
In particular through airborne weaponry, the government got the upper hand. 
The ``upsurge period'' from 1997 onwards also saw a rise of (anti-guerrilla) 
paramilitary activity whereas  before they had been involved in drug trafficking.  
The paramilitaries started operations around 1997.  
There was a consolidation of paramilitary 
groups in 1997, announced publicly in December
1997, and they became active then, having many of their own number killed.
It was not until 1999 that paramilitaries began to kill large numbers of guerrillas.
Among all of these mutually influencing, reinforcing or retarding, trends and 
events, our analysis points to a succession of two months where the global 
change was most intense.

We now come to the 2002 changepoint, (iii).  
A peak of (paramilitary) casualties brought about by government against 
paramilitaries was in 2002 due to aerial bombardment of a paramilitary 
position under attack by guerrillas.  It was a major setback for the 
paramilitaries, who declared a truce following the 
election that year of President 
Alvaro Uribe.

\section{Conclusions}

In our data mining examples we have shown how an ultrametric embedding is 
achieved, starting from the data.  The $\chi^2$ metric is appropriate for 
frequencies of occurrence data.  Points and associated masses in the 
dual spaces of observations and of attributes can be mapped from the $\chi^2$ 
distance to a Euclidean space.  On the points in this, now of identical 
masses, a hierarchy associated with an ultrametric can be induced.  
We have described briefly in this work how such a hierarchy can be used
to read off, and otherwise investigate, changepoints or anomalous  
occurrences at a range of scales.  

\section*{Appendix: the Correspondence Analysis and Hierarchical Clustering
Platform}

This Appendix introduces important aspects of Correspondence Analysis and 
hierarchical clustering.  Further reading is to be found in 
\cite{benz} and \cite{murtagh05}. 

\subsection*{Analysis Chain}

\begin{enumerate}
\item Starting point: a matrix that cross-tabulates the dependencies,
e.g.\ frequencies of joint occurrence, of an observations crossed by attributes
matrix.  
\item By endowing the cross-tabulation matrix with the $\chi^2$ metric 
on both observation set (rows) and attribute set (columns), we can map 
observations and attributes into the same space, endowed with the Euclidean
metric.  
\item A hierarchical clustering is induced on the Euclidean space, the 
factor space.  
\end{enumerate}

Various aspects of Correspondence Analysis follow on 
from this, such as Multiple Correspondence Analysis, different ways that
one can encode input data, and mutual description of clusters in terms of 
factors and vice versa.   In the following we use elements of 
the Einstein tensor notation of \cite{benz}.  This often reduces to common
vector notation.  

\subsection*{Correspondence Analysis: 
Mapping $\chi^2$ Distances into Euclidean Distances}

\begin{itemize}
\item The given contingency table (or numbers of occurrence) 
data is denoted $k_{IJ} =
\{ k_{IJ}(i,j) = k(i, j) ; i \in I, j \in J \}$.  

\item $I$ is the set of observation 
indexes, and $J$ is the set of attribute indexes.  We have
$k(i) = \sum_{j \in J} k(i, j)$.  Analogously $k(j)$ is defined,
and $k = \sum_{i \in I, j \in J} k(i,j)$.  

\item Relative frequencies: $f_{IJ} = \{ f_{ij}
= k(i,j)/k ; i \in I, j \in J\} \subset \R_{I \times J}$,
similarly $f_I$ is defined as  $\{f_i = k(i)/k ; i \in I, j \in J\}
\subset \R_I$, and $f_J$ analogously.  
\item The conditional distribution of $f_J$ knowing $i \in I$, also termed
the $j$th profile with coordinates indexed by the elements of $I$, is:

$$ f^i_J = \{ f^i_j = f_{ij}/f_i = (k_{ij}/k)/(k_i/k) ; f_i > 0 ;
j \in J \}$$ and likewise for $f^j_I$.  

\end{itemize}

\subsection*{Input: Cloud of Points Endowed with the Chi Squared Metric}

\begin{itemize}
\item The cloud of points consists of the couples: 
(multidimensional) profile coordinate and (scalar) mass.
We have $ N_J(I) = $
$\{ ( f^i_J, f_i ) ; i  \in I \} \subset \R_J $, and
again similarly for $N_I(J)$.  

\item Included in this expression is the fact
that the cloud of observations, $ N_J(I)$, is a subset of the real 
space of dimensionality $| J |$ where $| . |$ denotes cardinality 
of the attribute set, $J$.  

\item The overall inertia is as follows: 
$$M^2(N_J(I)) = M^2(N_I(J)) = \| f_{IJ} - f_I f_J \|^2_{f_I f_J} $$
$$= \sum_{i \in I, j \in J} (f_{ij} - f_i f_j)^2 / f_i f_j $$

\item The term  $\| f_{IJ} - f_I f_J \|^2_{f_I f_J}$ is the $\chi^2$ metric
between the probability distribution $f_{IJ}$ and the product of marginal
distributions $f_I f_J$, with as center of the metric the product
$f_I f_J$.  
\item Decomposing the moment of inertia of the cloud $N_J(I)$ -- or 
of $N_I(J)$ since both analyses are inherently related -- furnishes the 
principal axes of inertia, defined from a singular value decomposition.
\end{itemize}

\subsection*{Output: Cloud of Points Endowed with the Euclidean 
Metric in Factor Space}

\begin{itemize}

\item The $\chi^2$ distance with center $f_J$ between observations $i$ and 
$i'$ is written as follows in two different notations: 

$$
d(i,i') = \| f^i_J - f^{i'}_J \|^2_{f_J} = \sum_j \frac{1}{f_j} 
\left( \frac{f_{ij}}{f_i} - \frac{f_{i'j}}{f_{i'}} \right)^2
$$

\item In the factor space this pairwise distance is identical.  The coordinate
system and the metric change.  

\item For factors indexed by $\alpha$ and for 
total dimensionality $N$ ($ N = \mbox{ min } \{ |I| - 1, |J| - 1 \}$;
the subtraction of 1 is since 
the $\chi^2$ distance is centered and  
hence there is a linear dependency which 
reduces the inherent dimensionality by 1) we have the projection of 
observation $i$ on the $\alpha$th factor, $F_\alpha$, given by 
$F_\alpha(i)$: 

\begin{equation}
d(i,i') = \sum_{\alpha = 1..N} \left( F_\alpha(i) - F_\alpha(i') \right)^2
\end{equation}

\end{itemize}

\begin{itemize}

\item In Correspondence Analysis the factors are ordered by decreasing 
moments of inertia.  
\item The factors are closely related, mathematically, 
in the decomposition of the overall cloud, 
$N_J(I)$ and $N_I(J)$, inertias.  
\item The eigenvalues associated with the 
factors, identically in the space of observations indexed by set $I$, 
and in the space of attributes indexed by set $J$, are given by the 
eigenvalues associated with the decomposition of the inertia.  
\item The 
decomposition of the inertia is a 
principal axis decomposition, which is arrived at through a singular
value decomposition.  
\end{itemize}

\subsection*{Hierarchical Clustering}

Background on hierarchical clustering in general, and the 
particular algorithm used here, can be found in \cite{murtagh85}.

Consider the projection of observation $i$ onto the set of 
all factors indexed by $\alpha$, $\{ F_\alpha(i) \}$ for all $\alpha$,
which defines the observation $i$ in the new coordinate frame.  
This new factor space is 
endowed with the (unweighted) Euclidean distance,  $d$.  
We seek a hierarchical clustering that takes into account the 
observation  sequence,
i.e.\ observation $i$ precedes observation $i'$ for all $i, i' \in I$.  
We use the linear
order on the observation set.  
\smallskip

\noindent
Agglomerative hierarchical clustering algorithm:  

\begin{enumerate}
\item Consider each observation in the sequence as constituting a 
singleton cluster.  
Determine the closest pair of adjacent observations, and define a cluster
from them.
\item Determine and merge 
the closest pair of adjacent clusters, $c_1$ and $c_2$, 
where closeness is defined by $d(c_1, c_2) = \mbox{ max } \{ d_{ii'}$
$\mbox{ such that } i \in c_1, i' \in c_2 \}$.  
\item Repeat step 2 until only one cluster remains.
\end{enumerate}

This is a sequence-constrained complete link agglomeration criterion.  
The cluster proximity at each agglomeration is strictly non-decreasing.

\section*{Acknowledgements}

The media work is in collaboration with Adam Ganz and Stewart McKie,
Royal Holloway, University of London, Department of Media Arts.  The Colombia
work is in collaboration with Michael Spagat, Department of Economics.

\end{document}